\documentclass[10pt,twocolumn,letterpaper]{article}
\makeatletter
\@namedef{ver@everyshi.sty}{}
\makeatother
\usepackage{iccv}
\usepackage{times}
\usepackage{epsfig}
\usepackage{graphicx}
\usepackage{amsmath}
\usepackage{amssymb, bm}
\usepackage{booktabs}
\usepackage{multirow}
\usepackage{subfig}
\usepackage{tikz}
\usepackage{pgfplots}
\usepackage{url}
\usepackage{pifont}
\usepackage{caption}
\usepackage[accsupp]{axessibility}
\newcommand{\xmark}{\ding{55}}%
\graphicspath{{figures/}}

\usepackage[pagebackref=true,breaklinks=true,letterpaper=true,colorlinks,citecolor=blue, bookmarks=false]{hyperref}

\iccvfinalcopy 

\def\iccvPaperID{3565} 

\ificcvfinal\fi

\begin{document}

\title{Topic Scene Graph Generation by Attention Distillation from Caption}

\author{Wenbin Wang\textsuperscript{1,2}, 
Ruiping Wang\textsuperscript{1,2,3}, 
Xilin Chen\textsuperscript{1,2}\\
\textsuperscript{1}Key Laboratory of Intelligent Information Processing of Chinese Academy of Sciences (CAS),\\
Institute of Computing Technology, CAS, Beijing, 100190, China\\
\textsuperscript{2}University of Chinese Academy of Sciences, Beijing, 100049, China\\
\textsuperscript{3}Beijing Academy of Artificial Intelligence, Beijing, 100084, China\\
{\tt\small wenbin.wang@vipl.ict.ac.cn, \{wangruiping, xlchen\}@ict.ac.cn}\\
{\tt\small \url{https://github.com/Kenneth-Wong/MMSceneGraph}}
}

\maketitle
\ificcvfinal\thispagestyle{empty}\fi

\begin{abstract}
   If an image tells a story, the image caption is the briefest narrator. Generally, a scene graph prefers to be an omniscient ``generalist'', while the image caption is more willing to be a ``specialist'', which outlines the gist. Lots of previous studies have found that a scene graph is not as practical as expected unless it can reduce the trivial contents and noises. In this respect, the image caption is a good tutor. To this end, we let the scene graph borrow the ability from the image caption so that it can be a specialist on the basis of remaining all-around, resulting in the so-called \textbf{Topic Scene Graph}. What an image caption pays attention to is distilled and passed to the scene graph for estimating the importance of partial objects, relationships, and events. Specifically, during the caption generation, the attention about individual objects in each time step is collected, pooled, and assembled to obtain the attention about relationships, which serves as weak supervision for regularizing the estimated importance scores of relationships. In addition, as this attention distillation process provides an opportunity for combining the generation of image caption and scene graph together, we further transform the scene graph into linguistic form with rich and free-form expressions by sharing a single generation model with image caption. Experiments show that attention distillation brings significant improvements in mining important relationships without strong supervision, and the topic scene graph shows great potential in subsequent applications. 
\end{abstract}

\section{Introduction}

A picture is worth a thousand words. However, only a few person prefers to know all of the ``thousand words'', while others would like to be informed the ``topic words''. Therefore, the scene graph and the image caption are used for conveying the image contents out of different purposes.   
Concretely, the scene graph \cite{johnson2015image} consists of objects in an image and the relationships between pairs of objects. A series of studies have tried to generate the scene graph and realize its potential in advanced intelligence tasks, \eg, visual Q\&A \cite{antol2015vqa,Tang_2019_CVPR}, visual reasoning \cite{Shi_2019_CVPR}, and vision-and-language navigation (VLN) \cite{wang2019reinforced}, \etc. Nevertheless, as pointed in \cite{Li2019Know, milewski2020scene, wang2020sketching}, the scene graph is helpful only if it is informative, while the current generated scene graph with such a lot of noises does not meet this standard. This is mainly due to the explosive combination possibility of two objects \cite{wang2020sketching, yu2020visual}, which brings the double-edge effect that the scene graph is comprehensive but the key information is overwhelmed by massive trivial details. It is necessary and practical to make the scene graph well-circumscribed between important and trivial contents. Fortunately, the image caption exactly shows this ability and is a good teacher from which a scene graph should learn. 

\begin{figure}[t]
\setlength{\abovecaptionskip}{-0.2cm}
\setlength{\belowcaptionskip}{-0.4cm}
\begin{center}
\includegraphics[width=1.0\linewidth]{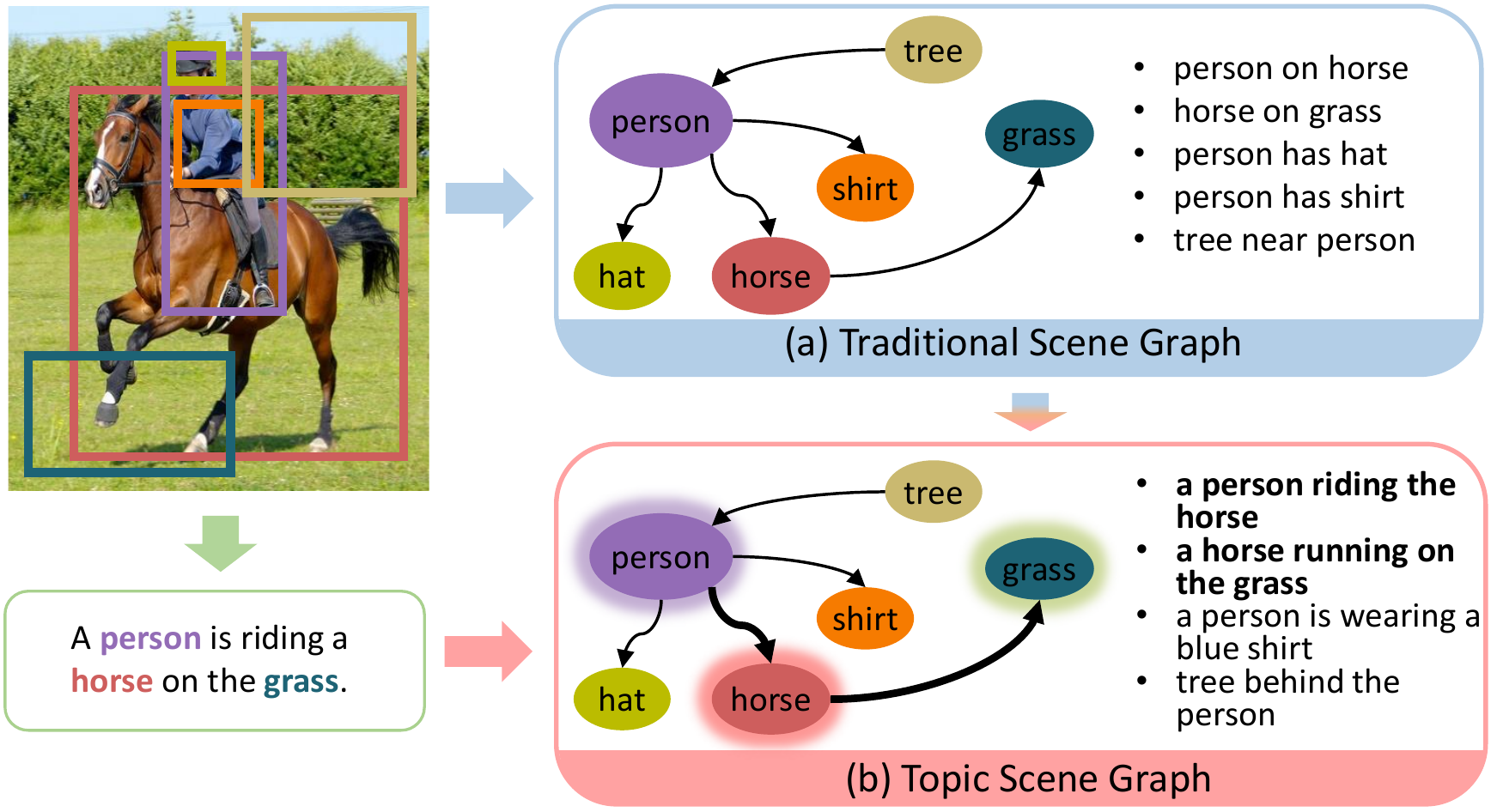}
\end{center}
   \caption{Comparison of the (a) traditional general scene graph, and the (b) topic scene graph generated under the guidance of attention in the image caption, which gives priority to the important relationships (highlighted nodes and edges), and expresses a relationship in the style of natural language. }
\label{fig:intro}
\end{figure}

In the context of scene graph generation, few researches devote endeavor to discovering the important relationships, which is a meaningful step for restricting the scale of the scene graph when it is used for downstream tasks.  
The most popular approach is to keep the relationships with large products of the predicted subject, object, and predicate scores. However, this product measures  the accuracy of prediction rather than the importance.   
Yang \etal \cite{yang2018graph} and Lv \etal \cite{lv2020avr} either use a light-weight relationship proposal network to extract some probably related pairs, or predict an attention score for each relationship, based on the perspective that annotated relationships are the important ones. This may be questionable because the mainstream scene graph datasets (\eg, Visual Genome \cite{krishna2017visual}) are suffered from serious long-tailed problem \cite{Chen_2019_CVPR,Tang_2020_CVPR} and the annotated pairs (head pairs) are usually trivial ones \cite{wang2020sketching}. To more precisely define what is important relationship, image caption is found helpful because a caption almost exactly reveals what humans think important \cite{He_2019_ICCV}, as shown in Figure \ref{fig:intro}. Consequently, Yu \etal \cite{yu2020visual} and Wang \etal \cite{wang2020sketching} learn to mine the important relationships under the strong supervision from the important relationship annotation which is obtained under the guidance of captions. But this process need to transform the captions into triplets first and then align two groups of heterologous triplets, which is so expensive and complicates the scene graph generation. 

In this work, we propose to let the scene graph learn the important relationships from the image caption in an economical way, resulting in the \textbf{Topic Scene Graph}. The importance of relationship is estimated by distilling the visual attention during the image caption generation, which is treated as the \textit{weak} supervision. Specifically, most advanced image captioners are able to fix its gaze on the correct object regions. We apply an image captioner and collect the \textit{first-order attention} information with respect to the object regions, which are used for assembling the \textit{second-order attention} about the relationships. In this way, we actually transform humans' attention into a new form, converting it from the concern about individual to that about relational events. The second-order attention is used as the weak supervision for guiding the estimation of the importance of the relationship. In this way, strong supervision is no longer necessary. 

Furthermore, as the attention distillation process makes it possible to generate the image caption and scene graph simultaneously and both of them are the description of an image, why not generating them with a single model? It is noted that the most popular scene graph dataset, Visual Genome (VG) \cite{krishna2017visual} of world scale contains more than 40,000 types of relationships which are originally extracted from humans' language, while the traditional definition of scene graph treats the relationship recognition as predicate classification and makes most of the relationships filtered, which is harmful to the diversity of relationship description. What is worse, there exist huge interior differences in some certain predicate classes, \eg, the appearance of two relationship triplet instances for the predicate class \textit{riding}, \textit{person-riding-horse} and \textit{dog-riding-skateboard} are totally different. It is difficult to clearly define the semantic boundaries between different predicates. 
Inspired by \cite{Kim_2019_CVPR}, we redefine the scene graph as the set of short relational sentences. In this way, a shared captioning module can be used for the so-called \textbf{linguistic scene graph} generation and image captioning at the same time. 

\section{Related Work}

\textbf{Scene Graph Generation (SGG) and Visual Relationship Detection (VRD)} focus on understanding the relationships between objects. Early studies \cite{divvala2014learning, sadeghi2011recognition} treat each distinct combination
of object categories and relationship predicates as a distinct
class. Lu \etal \cite{lu2016visual} formally define the VRD task and address the object and predicate classification separately. Recent state-of-the-art VRD works \cite{dai2017detecting,inayoshi2020bounding,li2017vip,mi2020hierarchical,peyre2017weakly,yin2018zoom,yu2017visual,zhang2017visual,zhang2017ppr,zhang2019large} pay attention on the prediction of each relationship triplet independently. The scene graph which describes the image faithfully from a bird's-eye view is proposed in \cite{johnson2015image}. After that, a batteries of studies contribute to generation of a high-quality scene graph. 
Message passing mechanism \cite{xu2017scene} has been proved effective and its variants are widely adopted in \cite{li2018factorizable, li2017scene}. The latest essential practice achieves more promising results through constructing reasonable context among objects and visual relationships \cite{Lin_2020_CVPR, Qi_2019_CVPR, Tang_2019_CVPR, Wang_2019_CVPR, yang2018graph, zellers2018neural}, or taking the advantage of external knowledge and commonsense \cite{Chen_2019_CVPR, Gu_2019_CVPR, zareian2020bridging, zareian2020learning}. Besides, 
Zhang \etal \cite{Zhang2019graphical} propose contrastive losses to resolve the related pair configuration ambiguity. Zareian \etal \cite{Zareian_2020_CVPR} creatively consider SGG as an edge role assignment problem. Tang \etal \cite{Tang_2020_CVPR} diversify the predicted relationships through addressing the causal effect. Most of these works struggle to fit the VG dataset but always overlook the fact that the scene graph annotation suffers from serious long-tailed problem and the valuable relationships are usually overwhelmed by trivial ones. This problem is general because of the reporting bias \cite{gordon2013reporting} and should not be blamed on a particular dataset. A growing number of works are considering how to make the scene graph more practical. Liang \etal \cite{liang2019vrr} prune the dominant and easy-to-predict relations while keeping the visually relevant relations in VG. Lv \etal \cite{lv2020avr} estimate the importance of relationships with an attention module, but actually they still think that the annotated relationships are semantically important, which may not be true. Yu \etal \cite{yu2020visual} and Wang \etal \cite{wang2020sketching} provide annotations with relationships of humans interest under guidance of image caption and explicitly use them as supervision. However, only semantically important relationships are detected in \cite{yu2020visual}, which is not enough for a comprehensive scene graph. In this work, we distill the attention from image caption as weak supervision rather than constructing the high-cost annotation of important relationships, and reasonably estimate the importance of all relationships so that the number of remaining relationships are controllable.

\textbf{Image captioning and scene graph}.  
Compared to the scene graph, the image caption is usually treated as the final presentation to humans (for interaction). Restricted by the length, an image caption usually contains the most important contents in an image \cite{He_2019_ICCV}, but misses details. Some researchers propose the dense captioning task \cite{johnson2016densecap} which generates diverse but aimless region-level descriptions passively. Our proposed topic scene graph is naturally a structured representation of an image, and is  especially able to actively estimate the importance of the image contents.

Early studies on image captioning are rule or template-based \cite{socher2010connecting, yao2010i2t}. Modern captioning models have achieved great progress benefiting from encoder-decoder  framework \cite{vinyals2015show}, attention technique \cite{ chen2017sca,cornia2020meshed,herdade2019image,huang2019attention,li2019entangled,pan2020x,vaswani2017attention,xu2015show,yang2016stacked} and RL-based training objective \cite{rennie2017self}. Our work distills the attention from image caption. Besides, as the scene graph contains much semantic information, lots of works have tried to incorporate it into captioning models \cite{chen2020say,Gu_2019_ICCV,Li2019Know,milewski2020scene,xu2019scene,Yang_2019_CVPR,yao2018exploring, yao2019hierarchy}.  Inspired by this direction, we propose to generate linguistic scene graph, and creatively let the scene graph benefit from image caption.

\begin{figure*}[t]
\setlength{\abovecaptionskip}{-0.2cm}
\setlength{\belowcaptionskip}{-0.4cm}
\begin{center}
\includegraphics[width=1.0\linewidth]{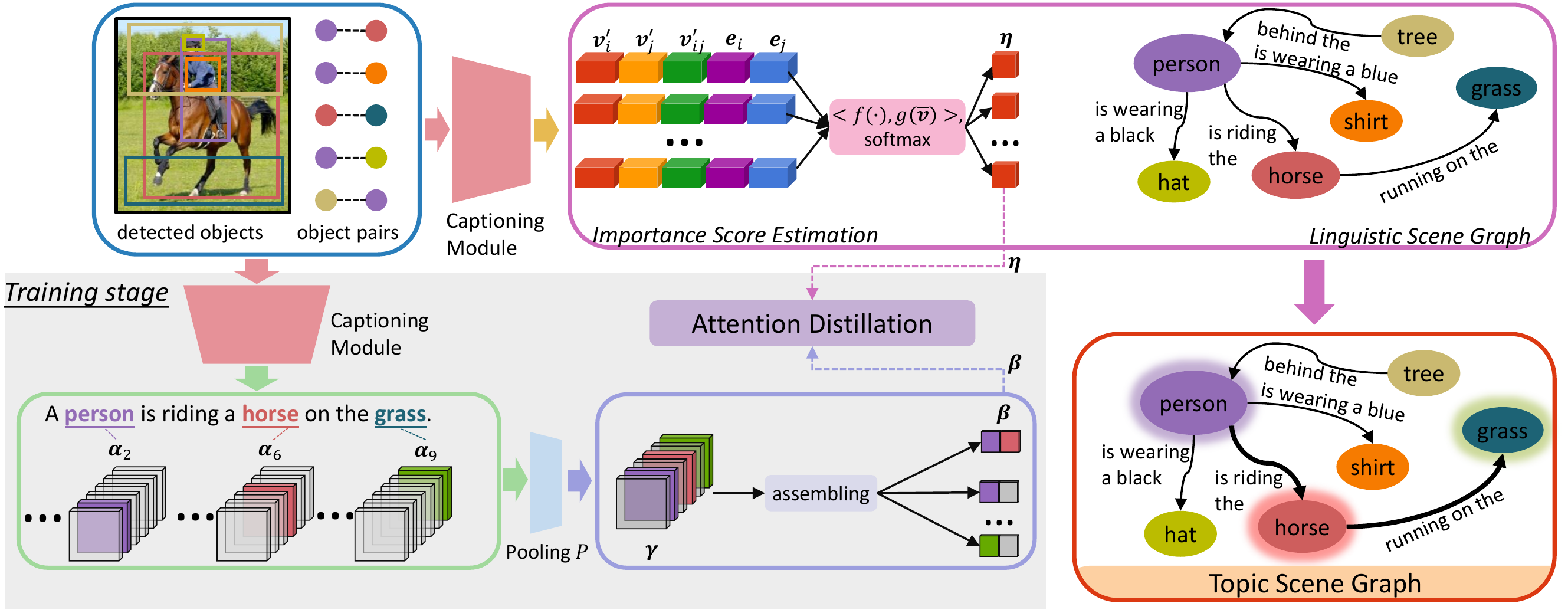}
\end{center}
   \caption{The framework of our method. The detected objects and the object pairs are fed into a shared captioning module to generate the image caption and relational captions which form a linguistic scene graph. During the training stage, the attention information is collected, pooled over multiple time steps, and assembled to produce the second-order attention $\bm{\beta}$ about the relationships. Simultaneously the importance scores $\bm{\eta}$ of the relationships are also estimated, and then regularized by $\bm{\beta}$. In the testing phase, the estimated importance scores are used for generating the final topic scene graph. }
\label{fig:framework}
\end{figure*}

\section{Approach}
\subsection{Overview}
Given an image $I$, its scene graph $G$ consists of a set of objects (nodes) $O=\{o_i \mid i\in[1, n]\}$ with the assigned class labels $C=\{c_i \mid i\in [1, n], c_i\in \mathcal{C}\}$ and the corresponding bounding boxes $B=\{b_i\mid i\in [1, n], b_i \in \mathbb{R}^4\}$, and a set of relationships (edges) $R=\{r_k\mid k\in [1, m]\}$. Conventionally, each relationship $r_k$ is a triplet of the start node $o_i$, the end node $o_j$, and the relationship label $x_{ij}\in \mathcal{R}$ where $\mathcal{R}$ is the set of predicate types. These relationships are disordered. Thinking of the limitation of this representation, we redefine the relationship $r_k$ as a relational caption in the form of word sequence $(y_{kt})_{t=1}^{T_R}$, where $y_{kt} \in \mathcal{V}$ and $\mathcal{V}$ denotes the vocabulary. $t$ is the positional index of the word in the sequence and $T_R$ is the sequence length. More importantly, these relationships are sorted according to their importance. Specifically, as depicted in Figure \ref{fig:intro} (b), the detected objects are used for generating the relational captions and the image caption $(w_t)_{t=1}^{T_C}$ ($w_t\in \mathcal{V}$ and $T_C$ is the caption length), during which the subjective interest (attention) $\bm{\alpha}$ is collected from the image caption. The $\bm{\alpha}$ is used for the estimation of importance scores of the relationships. 



In the following parts, we will describe the generation of the image caption (Sec. \ref{sec:cap}) and the linguistic scene graph (Sec. \ref{sec:lingsg}) using a shared captioning module. Then we will elaborate on importance score estimation and attention distillation for obtaining the final topic scene graph (Sec. \ref{sec:tsg}). 
 
 \subsection{Captioning Module}\label{sec:cap}
In this work, we adopt two types of state-of-the-art captioning models, the Up-Down model \cite{anderson2018bottom}  based on LSTM \cite{hochreiter1997long}, and  the Transformer \cite{vaswani2017attention}. The Up-Down model comprises of an attention LSTM layer and a language LSTM layer. Specifically, the detected objects are represented by their visual features $\bm{V}=[\bm{v}_1, \ldots, \bm{v}_n]\in \mathbb{R}^{d_v \times n}$ and bounding boxes $[b_1, \ldots, b_n]\in \mathbb{R}^{4\times n}$. The object visual features are firstly transformed to $\bm{V}'=[\bm{v}'_1, \ldots, \bm{v}'_n]\in \mathbb{R}^{d_l \times n}$ with a lower dimension: 
\begin{equation}
    \bm{v}'_i = \bm{W}_v \bm{v}_i + \bm{b}_v, \quad \bm{v}'_i \in \mathbb{R}^{d_l},
\end{equation}
where $\bm{W}_v\in \mathbb{R}^{d_l\times d_v}$ and $\bm{b}_v\in \mathbb{R}^{d_l}$ are trainable parameters. At each time step $t$, the previous hidden state of the language LSTM $\bm{h}^2_{t-1}$ is concatenated with the mean-pooled image feature $\overline{\bm{v}}=\frac{1}{n} \sum_i \bm{v}'_i$ and the previous word embedding $\bm{e}_{t-1}=\bm{W}_e w_{t-1}$, and fed into the attention LSTM:
\begin{equation}
    \bm{h}^1_t = \mathrm{LSTM}_{\mathrm{Att}}([\bm{h}^2_{t-1}; \overline{\bm{v}}; \bm{e}_{t-1}], \bm{h}^1_{t-1}), \quad \bm{h}^1_t\in \mathbb{R}^{d_h},
\end{equation}
where $[;]$ denotes the concatenation and $\bm{W}_e \in \mathbb{R}^{d_e \times |\mathcal{V}|}$ is the embedding matrix. The $w_{t-1}$ stands for the $|\mathcal{V}|$-dim one-hot vector where the $w_{t-1}$-th element is 1 in practice.  The attention about the objects $\bm{\alpha}=[\bm{\alpha}_1, \ldots, \bm{\alpha}_{T_C}] \in \mathbb{R}^{n\times T_C}$ is calculated as:
\begin{equation}
\begin{aligned}
z_{i,t} &= \bm{w}_a^\mathrm{T}\tanh{(\bm{W}_{va}\bm{v}'_i + \bm{W}_{ha}\bm{h}^1_t)},\\
\bm{\alpha}_t &= \mathrm{softmax}(\bm{z}_t),
\end{aligned}
\end{equation}
where $\bm{w}_a\in \mathbb{R}^{d_a\times 1}$, $\bm{W}_{va}\in \mathbb{R}^{d_a\times d_l}$, and $\bm{W}_{ha}\in \mathbb{R}^{d_a\times d_h}$ are trainable parameters. 
Finally, the attended image visual feature $\hat{\bm{v}}_t=\sum_{i=1}^n z_{i,t} \bm{v}'_i$  and $\bm{h}^1_t$ are used as the input of the language LSTM, which predicts the conditional distribution over the possible word:
\begin{equation}
\begin{aligned}
    \bm{h}^2_t=\mathrm{LSTM}_{\mathrm{Lang}}([\hat{\bm{v}}_t;\bm{h}^1_t], \bm{h}^2_{t-1}), \quad \bm{h}^2_t \in \mathbb{R}^{d_h},\\
    p(w_t|w_{1:t-1}) = \mathrm{softmax}(\bm{W}_o\bm{h}^2_t + \bm{b}_o),
\end{aligned}
\end{equation}
with trainable parameters $\bm{W}_o\in \mathbb{R}^{|\mathcal{V}|\times d_h}$ and $\bm{b}_o\in \mathbb{R}^{|\mathcal{V}|}$. 

As for the Transformer model, it consists of an encoder and a decoder, both of which contain a stack of layers. We provide details in the Supp. and especially explain how to extract the attention on objects here. For captioning task, the transformed visual features $\bm{V}'$ are fed into the encoder and we get the output $\bm{V}^*\in \mathbb{R}^{d_{tr}\times n}$. For each decoder layer in the decoder, it contains a multi-head self-attention layer and a multi-head cross-attention layer. All the word embeddings $\bm{E} $ are fed into the self-attention layer to get the output $\bm{E}^* \in \mathbb{R}^{
d_{tr}\times T_C}$. In each head $j\in [1, H]$, the attention weights $\bm{\alpha}^j\in \mathbb{R}^{n\times T_C}$ about objects are computed by:
\begin{equation}
    \bm{\alpha}^j = \mathrm{softmax}\left(\frac{{\bm{V}^*}^\mathrm{T}{\bm{E}^*}}{\sqrt{d_{tr}}}\right).
\end{equation}
We average the $\bm{\alpha}^j$ across $H$ heads and obtain the final $\bm{\alpha}$. 

\subsection{Linguistic Scene Graph}\label{sec:lingsg}
In this part, we share the captioning module to make it applicable to relational captioning so that a linguistic scene graph is realizable. Following the general scene graph generation process, we build combinations of $N$ detected objects and obtain $O(N^2)$ object pairs. For the subject $o_i$ and the object $o_j$, we extract their union visual feature $\bm{v}_{ij}\in \mathbb{R}^{d_u}$ which contains rich context information by applying ROI pooling \cite{ren2015faster} with the union box of $b_i$ and $b_j$. Besides, as the relative position between two objects is found as effective prior information, we follow \cite{peyre2017weakly} to build the geometry feature:
\begin{equation}
    \bm{g}_{ij} =\!\!\left[ \frac{x_j - x_i}{\sqrt{w_i h_i}},\! \frac{y_j-y_i}{\sqrt{w_i h_i}}, \!\sqrt{\frac{w_j h_j}{w_i h_i}},\! \frac{w_i}{h_i},\! \frac{w_j}{h_j}, \!\frac{b_i\cap b_j}{b_i \cup b_j}\right],
\end{equation}
where the $(x, y)$ is the center position and $w$ and $h$ denote width and height of a box. It is further projected to a 64-dim feature and concatenated with $\bm{v}_{ij}$ to obtain  
the final union feature $\bm{v}'_{ij} \in \mathbb{R}^{d_l}$:
\begin{equation}
    \bm{v}'_{ij} = \bm{W}_u[\bm{v}_{ij}; (\bm{W}_g \bm{g}_{ij}+\bm{b}_g)]+\bm{b}_u,
\end{equation}
where $\bm{W}_u\in \mathbb{R}^{d_l \times (d_u+64)}$, $\bm{b}_u\in \mathbb{R}^{d_l}$, $\bm{W}_g\in \mathbb{R}^{64\times 6}$, and $\bm{b}_g\in \mathbb{R}^{64}$ are trainable parameters. 


Different from the image captioning which should pay attention to all objects, relational captioning focuses on two designated objects. Specifically, for the Up-Down model, only the $\bm{v}'_i$, $\bm{v}'_j$ and $\bm{v}'_{ij}$ are used for decoding. For the Transformer, all object features are fed into the encoder to construct contextual information $\bm{V}^*$, but only the $\bm{v}^*_i$, $\bm{v}^*_j$ and $\bm{v}'_{ij}$ are fed into the decoder.

\subsection{Topic Scene Graph}\label{sec:tsg}
With the attention about objects provided by image captioning, we propose to assemble it to obtain attention about the relationships, which is used as the weak supervision to guide the estimation of the importance of the relationships.  

Suppose that there are $m$ relationships. We first estimate the importance score for each relationship consisting of subject $o_i$ and object $o_j$. Specifically, as depicted in the top middle part in Figure \ref{fig:framework}, we concatenate the $\bm{v}'_i$, $\bm{v}'_j$, $\bm{v}'_{ij}$ and the semantic embeddings of the subject and object categories, $\bm{e}_i$, $\bm{e}_j \in \mathbb{R}^{d_{sem}}$, to form a query $\bm{q}_{ij}$, and compute the key using the global feature $\overline{\bm{v}}$:
\begin{align}
    \bm{q}_{ij} &= f([\bm{v}'_i; \bm{v}'_j; \bm{v}'_{ij}; \bm{e}_i; \bm{e}_j]) \in \mathbb{R}^{d_s}, \\
    \bm{k}&=g(\overline{\bm{v}}) \in \mathbb{R}^{d_s},
\end{align}
where $f$ and $g$ are two learnable linear transformation functions. The estimated importance score $s_{ij}$ is calculated as the inner product of the query and key and then normalized with softmax function:
\begin{align}
    s_{ij}& = \frac{\bm{q}_{ij}^\mathrm{T}\bm{k}}{\sqrt{d_s}},\\
    \bm{\eta}&=\mathrm{softmax}(\bm{s}) \in \mathbb{R}^m.
\end{align}

On the other hand, we have the attention information $\bm{\alpha}$ with respect to individual objects, which is used to assemble the attention $\bm{\beta}$ with respect to relationships. As shown in the bottom left part (with gray background) in Figure \ref{fig:framework}, firstly, we gather the attention score for each object over multiple time steps with a pooling function $P$, resulting in $\bm{\gamma}\in \mathbb{R}^n$. Then the second order attention for a relationship is assembled as:
\begin{equation}
    \delta_{ij}=\gamma_i + \gamma_j, \quad 
    \bm{\beta} =\mathrm{softmax}(\bm{\delta}) \in \mathbb{R}^m.
\end{equation}
Finally, the estimated score $\bm{\eta}$ is regularized with the induced second-order attention $\bm{\beta}$ via KL-divergence. 

\subsection{Optimization}
The optimization process is divided into two stages. In the first stage, given a single ground truth image caption $(w_t)_{t=1}^{T_C}$ and $m$ ground truth relational captions $[(y_{kt})_{t=1}^{T_R}]_{k=1}^m$, the captioning module is optimized with the traditional cross-entropy loss consisting of the image captioning part and the relational captioning part:
\begin{equation}
\begin{split}
    L_{CE}&=\sum_{t=1}^{T_C} -\log p(w_t|w_{1:t-1}) \\
    &+ \lambda \left(\frac{1}{m}\sum_{k=1}^m\sum_{t=1}^{T_R} -\log p\left(y_{kt}|y_{k1:k(t-1)}\right)\right),
\end{split}
\end{equation}
where $\lambda$ is the balance parameter.  
In the second stage, the attention distillation module is optimized with the KL-divergence loss:
\begin{equation}
    L_{KL}=KL(\bm{\eta}||\bm{\beta}).
\end{equation}
Although the reinforce algorithm such as SCST \cite{rennie2017self} is widely used for further optimizing the captioning models, some researches \cite{zhou2020more} found that SCST actually does harm to the text-to-image grounding because it encourages the n-gram consistency rather than visual semantic alignment. As our framework has a high demand for superior grounding performance, we do not optimize the captioning module with SCST in this research and leave it to future works. 

\section{Experiments}
\subsection{Datasets and Evaluation Metrics}
\textbf{Datasets.} There is no existing dataset with both the image caption and relational caption. Inspired by \cite{Kim_2019_CVPR}, we refer to their data construction procedure to collect the relational caption, and further collect the image caption, as well as the important relationship annotation which is only for training the upper bound models and evaluation. Specifically, we use the 51,208 images in both VG and MSCOCO \cite{lin2014microsoft} datasets which have both relationship and image caption annotation. Firstly, we cleanse the VG dataset and keep a large scale vocabulary including 3,000 object categories and 800 attributes. The relationships about the objects beyond these categories are filtered. To obtain the important relationship annotation, we apply the Scene Graph Parser \cite{schuster2015generating} to extract the relationships from the image caption, and align them with the annotated ones by matching their subject and object WordNet \cite{Miller1992WORDNETAL} synsets. 
Finally we convert the remaining relationships into sentences. It is worth mentioning that we do not make a category-wise vocabulary for relationships, but keep the relationships in their original free and open form. To further enrich the concepts, the attributes of each object are randomly selected to add to the sentences. We obtain 35,928 images for 29,928/1,000/5,000 splits for train/validation/test sets respectively and 11,437 vocabularies (including 3,000 object categories and 800 attributes). 

\textbf{Evaluation metrics.} We use BLEU, METEOR, CIDEr-D, ROUGE-L, and SPICE for image captioning. For relational captioning, we refer to \cite{johnson2016densecap} and \cite{Kim_2019_CVPR} and use the following metrics. (1) mean Average Precision (mAP): it uses METEOR score \cite{denkowski2014meteor} with thresholds $\{0, 0.05, 0.10, 0.15, 0.20, 0.25\}$ for language and IoU thresholds $\{0.2, 0.3, 0.4, 0.5, 0.6\}$ for localization. Only the pair whose subject and object have IoUs greater than thresholds is a true positive sample. The mAP is obtained by averaging across all the combinations of language and localization thresholds. (2) image-level recall (Img-Lv.Recall): it ignores the localization and evaluate the recall of the bag of predicted relational captions. 

Besides, in order to evaluate whether the important relationships are properly found, we refer to the metrics in traditional scene graph generation \cite{ wang2020sketching,xu2017scene}, \ie, Recall@K where K is set to 20, 50, and 100. Under this metric, only the important relationships are regarded as ground truth and the top K relationships are evaluated, which means that the predicted relationships should be sorted. A relational caption is correct only if the following two conditions are satisfied: (1) both the subject and object have IoUs greater than 0.5, and (2) the METEOR score is greater than the thresholds above. We average the recall on different language thresholds. To evaluate the performance on discovering correct important object pairs, we derive the Recall-ns@K metric which only requires the above first condition and does not consider the METEOR scores.

\subsection{Implementation Details}
We firstly train the faster-RCNN \cite{ren2015faster} detector with the ResNeXt-101 \cite{Xie2017Aggregated} backbone on the objects of 3,000 categories of our dataset. During the scene graph training, the parameters of the object detector are frozen. More details are given in the Supp.

\subsection{Experiments on Linguistic Scene Graph}
In our method, a shared captioning module is trained for image captioning (IC) and relational captioning (RC), which has never been explored before. We start with exploring the effectiveness of this practice. To this end, we adjust the balance parameter $\lambda$ to control the proportion of the loss of the relational captioning in the final loss function . Results under more value settings are provided in Supp. The evaluation is divided into two parts. On the side of image captioning, the baselines are Up-Down (UD) and Transformer, which are trained only using the image captions. We compare the baselines to the UD-ICRC and Transformer-ICRC trained with the image captions and relational captions. From Table \ref{tab:joint-evalcap}, we observe that mixed training actually brings benefit to the image captioning, but as the $\lambda$ increases, this benefit will slightly drop. It suggests that mixed training is feasible despite the fact that the assembled relational captions will bring some noises. 

On the side of relational captioning, we use TriLSTM \cite{Kim_2019_CVPR}, UD-RC, and Transformer-RC as baselines, which are trained only using the relational captions. The TriLSTM is re-implemented and trained on our dataset. The relational captions are sorted by the product of the probabilities of the generated words, \ie, likelihoods. As shown in Table \ref{tab:joint-evalrc}, compared to the TriLSTM, both the UD-RC and UD-ICRC outperform it obviously. 
Comparing the UD-RC with UD-ICRC, we find that as the $\lambda$ increases, the UD-ICRC roughly performs better on the image level metrics, and surpasses the UD-RC baseline when $\lambda$ is greater than 0.7. However, the performance drops on the important relationship recall metrics. We think that it is because the increasing $\lambda$ makes the model fit the relational captions data better, but the increasing sentence likelihood loses its discrimination and is less suitable for importance estimation. It also suggests that neither the sentence likelihood of the relational caption nor the score product of the traditional triplet are unstable for importance estimation. As for the Transformer, mixed training has little impact on the performances.   With a comprehensive consideration on the performances of the two tasks, we set the $\lambda$ as 0.7 and freeze the UD-ICRC / Transformer-ICRC models for generating the topic scene graph in the following experiments. 

\begin{table}[]
\setlength{\abovecaptionskip}{-0.2cm}
\caption{Image captioning results. B1, B4, M, R, C, S denote BLEU-1, BLEU-4, METEOR, ROUGE-L, CIDEr-D, and SPICE respectively. ``-ICRC'' denotes that the model is trained with image captions and relational captions.}
\begin{center}
\resizebox{0.46\textwidth}{!}{
\begin{tabular}{@{}c|c|cccccc@{}}
\hline
Model                    & $\lambda$ & B1            & B4            & M             & R             & C             & S                       \\ \hline
UD \cite{anderson2018bottom}                       & -         & 69.8          & 29.6          & 25.0          & 52.3          & 94.1          & 18.0                    \\
\multirow{4}{*}{UD-ICRC} & 0.1       & \textbf{71.1} & \textbf{30.4} & \textbf{25.1} & \textbf{52.6} & \textbf{95.3} & \textbf{18.3}  \\
                         & 0.3       & 70.7          & 30.0          & 24.9          & 52.5          & 94.6          & 18.1                    \\
                         & 0.7       & 70.5          & 30.1          & 25.0          & 52.4          & 94.8          & 18.2                   \\
                         & 1.0       & 71.0          & 30.0          & 24.8          & 52.5          & 93.5          & 17.9                   \\ \hline
Transformer \cite{vaswani2017attention}              & -         & 68.8          & 26.8          & 23.5          & 50.4          & 85.6          & 17.3                    \\
Transformer-ICRC         & 0.7       & \textbf{70.3} & \textbf{28.6} & \textbf{24.4} & \textbf{51.7} & \textbf{91.5} & \textbf{18.0} \\ \hline
\end{tabular}}
\vspace{-0.8cm}
\end{center}
\label{tab:joint-evalcap}
\end{table}

\begin{table*}[]
\setlength{\abovecaptionskip}{-0.3cm}
\caption{Results of relational captioning (\%). ``-RC'' denotes that the model is only trained with the relational captions. ``-ICRC'' denotes that the model is trained with image captions and relational captions. R-ns means Recall-ns. Img-Lv. Recall means the image level recall. }
\begin{center}
\resizebox{0.9\textwidth}{!}{
\begin{tabular}{@{}c|c|ccccccccc@{}}
\hline
Model                    & $\lambda$            & mAP                  & METEOR               & Img-Lv. Recall       & R@20                 & R-ns@20              & R@50                 & R-ns@50              & R@100                & R-ns@100             \\ \hline
TriLSTM \cite{Kim_2019_CVPR}                 & -                    & 3.80                 & 30.21                & 72.72                & 1.31                 & 3.20                 & 3.93                 & 9.58                 & 8.42                 & 20.88                \\ \hline
UD-RC \cite{anderson2018bottom}                   & -                    & \textbf{5.61}        & 42.40                & 88.77                & 3.02                 & 3.71                 & \textbf{10.46}       & 12.92                & \textbf{22.97}       & 28.90                \\
\multirow{4}{*}{UD-ICRC} & 0.1                  & 4.84                 & 38.31                & 84.81                & \textbf{3.45}        & \textbf{4.43}        & 10.22                & \textbf{13.99}       & 20.77                & \textbf{29.00}       \\
                         & 0.3                  & 5.14                 & 40.36                & 86.93                & 3.39                 & 4.18                 & 9.87                 & 12.88                & 21.57                & 27.99                \\
                         & 0.7                  & 5.43                 & 42.26                & 89.15                & 2.75                 & 3.49                 & 9.97                 & 12.40                & 20.76                & 26.46                \\
                         & 1.0                  & 5.41                 & \textbf{42.75}       & \textbf{89.52}       & 2.31                 & 2.90                 & 8.09                 & 10.20                & 19.97                & 25.56                \\ \hline
Transformer-RC \cite{vaswani2017attention}             &                      & 5.26                 & 41.62                & 88.65                & 2.11                 & 2.73                 & 6.83                 & 9.12                 & 16.36                & 21.91                \\
Transformer-ICRC         & 0.7                  & 5.15                 & 41.63                & 88.64                & 2.05                 & 2.70                 & 6.86                 & 9.19                 & 16.21                & 21.91                \\ \hline
\end{tabular}}
\end{center}
\label{tab:joint-evalrc}
\end{table*}

\begin{table*}[]
\setlength{\abovecaptionskip}{-0.2cm}
\setlength{\belowcaptionskip}{-0.4cm}
\caption{Results (\%) comparison on discovering the important relationships. ``Feat.'' denotes different input features. ``P'' denotes the pooling function. ``Mask'' denotes masking the non-noun words (\checkmark) or not (\xmark).}
\begin{center}
\resizebox{0.98\textwidth}{!}{
\begin{tabular}{@{}c|ccc|ccccccc@{}}
\hline
                               & Feat. & $P$  & Mask                      & R@20           & R-ns@20        & R@50           & R-ns@50        & R@100          & R-ns@100       & mean           \\ \hline
TriLSTM                        & -     & -    & -                         & 1.31           & 3.20           & 3.93           & 9.58           & 8.42           & 20.88          & 7.89           \\ \hline
UD-ICRC                        & -     & -    & -                         & 2.75           & 3.49           & 9.97           & 12.40           & 20.76          & 26.46          & 12.64          \\ \hline
\multirow{6}{*}{UD-ICRC-attn}  & U     & MAX  & \xmark     & 7.27           & 10.53          & 17.12          & 24.10           & 30.44          & 42.22          & 21.95          \\
                               & SO    & MAX  & \xmark     & 7.49           & 10.88          & 20.61          & 28.79          & 37.06          & 51.07          & 25.98          \\
                               & SOU   & MAX  & \xmark     & \textbf{15.71} & 21.80          & 28.85          & 39.39          & 41.09          & \textbf{55.73} & 33.76          \\
                               & SOUS  & MEAN & \checkmark & 2.74           & 4.53           & 8.76           & 13.71          & 19.27          & 28.05          & 12.84          \\
                               & SOUS  & MAX  & \checkmark & 10.72          & 15.43          & 21.59          & 30.26          & 34.43          & 47.34          & 26.63          \\
                               & SOUS  & MAX  & \xmark     & 15.46          & \textbf{21.81} & \textbf{29.55} & \textbf{40.72} & \textbf{41.14} & 55.68          & \textbf{34.06} \\ \hline
\multirow{4}{*}{UD-ICRC-label} & U     & -    & -                         & 13.04          & 17.35          & 25.25          & 33.28          & 36.72          & 49.22          & 29.14          \\
                               & SO    & -    & -                         & 30.14          & 38.86          & 41.45          & 53.95          & 51.55          & 67.70          & 47.28          \\
                               & SOU   & -    & -                         & 32.17          & 41.38          & 43.57          & 56.68          & 53.65          & 70.81          & 49.71          \\
                               & SOUS  & -    & -                         & \textbf{34.39} & \textbf{45.13} & \textbf{46.03} & \textbf{60.97} & \textbf{54.60}  & \textbf{72.44} & \textbf{52.26} \\ \hline
Transformer-ICRC               & -     & -    & -                         & 2.05           & 2.70           & 6.86           & 9.19           & 16.21          & 21.91          & 9.82           \\
Transformer-ICRC-attn          & SOUS  & MAX  & \xmark     & \textbf{17.52} & \textbf{24.96} & \textbf{31.88} & \textbf{44.46} & \textbf{43.71} & \textbf{61.10} & \textbf{37.27} \\ \hline
Transformer-ICRC-label         & SOUS  & -    & -                         & \textbf{25.79} & \textbf{34.68} & \textbf{39.06} & \textbf{53.02} & \textbf{48.76} & \textbf{66.43} & \textbf{44.62} \\ \hline

\end{tabular}}
\end{center}
\label{tab:attn}
\end{table*}

\subsection{Experiments on Topic Scene Graph}
As we know this is the first time to study the topic linguistic scene graph generation. We replace some key components to show the effectiveness of our proposed model (UD-ICRC-attn) and facilitate the ablation study. As we have the important relationship annotation, we train the upper bound models named as UD-ICRC-label and Transformer-ICRC-label under the supervision of the annotated important relationships with  binary cross entropy loss.  The results are shown in Table \ref{tab:attn}. 

\textbf{Pooling function}. The pooling function $P$ is used for gathering attention information over multiple time steps for each individual object. We compare two functions: max pooling (MAX) and mean pooling (MEAN). Comparing the 4th row and the 5th row in the UD-ICRC-attn section, the max pooling function is much more effective than mean pooling. It is reasonable because we want to maximize the scores of the objects which are mentioned in the image captions, while the mean pooling reduces the attention scores and makes it hard to shed light on the key objects. 

\textbf{Input features}. We try to use different concatenated features for obtaining the query $\bm{q}$ when estimating the importance scores, including the union features (U), subject and object features (SO), subject, object, and union features (SOU), and subject, object, union features together with the semantic embeddings of subject and object categories (SOUS). By comparing the 1st$\sim$3rd rows and the 6th row in the UD-ICRC-attn section, and the rows in the UD-ICRC-label section, it is found that the SOU significantly improves the performances compared with U and SO, suggesting that these three types of features cannot be used independently, as the SO provides information about objects and the U provides relative spatial information. The semantic embeddings bring slight improvement, and it is not as obvious as that in the upper bound models.

\textbf{Masking non-noun words}. When gathering the attention information, we explore whether all the words of a sentence should be considered or not. Different from considering all words, we try an alternative way that only the attention of noun words are collected since they are probably to be correctly grounded to the regions, and other words are masked. To this end, we apply the NLTK POS tagger \cite{bird2006nltk} to filter out the non-noun words. Comparing  the 5th row and 6th row in the UD-ICRC-attn section,  it is interesting to find that masking the non-noun words does harm to the performance instead. This phenomenon may imply that the context plays a crucial role and the non-noun words would also contribute to the attention of the center nouns.

 Overall, compared with the TriLSTM, UD-ICRC and Transformer-ICRC baselines, 
 the application of our attention alignment module significantly improves the performances, and obviously reduces the gap between the baselines and the upper bound. The best configuration is to use the SOUS input features, max pooling function and collect the attention from all words. It's noted that our method does not need the complicated collection of important relationship annotation, but can still provide the useful important relationships. We observe that the attention alignment module is more effective for the Transformer, which may imply that the attention in Transformer is more precise.
 
 \begin{figure*}[t]
 \setlength{\abovecaptionskip}{-0.2cm}
 \setlength{\belowcaptionskip}{-0.1cm}
\begin{center}
\includegraphics[width=1.0\linewidth]{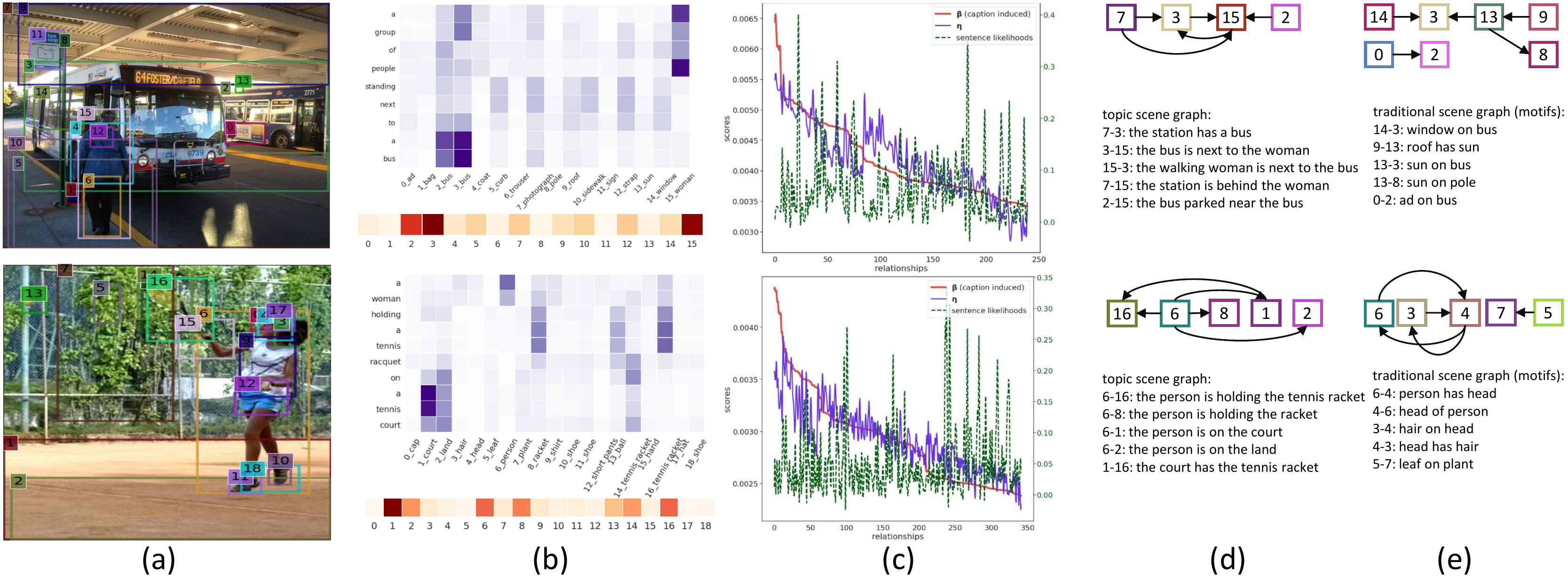}
\end{center}
   \caption{The qualitative results. (a) The objects with their bounding boxes and ids. (b) The attention about the objects during caption generation and the pooled attention are visualized. Darker colors indicate larger weights.  (c) Importance scores of the relationships are drawn. Along the X-axis, the relationships are sorted by the $\bm{\beta}$ scores in a descending order. All the lines are smoothed. (d-e) The scene graph from our method and motifs \cite{zellers2018neural} consisting of top 5 relationships are shown.}
\label{fig:visattn}
\end{figure*}

\subsection{Qualitative Results}
In Figure \ref{fig:visattn} (b), we visualize the attention about the objects of each word (the purple heat map) during captioning and the pooled attention over all words (the reddish brown heat map). It can be observed that although the caption may be not so precise, the objects are still correctly attended (the \textit{15\_woman} and \textit{3\_bus} in the first sample, the \textit{1\_court}, \textit{6\_person}, and the \textit{16\_tennis racket} in the second sample). The max pooling function highlights the mentioned objects in the caption. It's also found that an object can be activated by several words, which gives an explanation for the performance drop when masking the non-noun words. In Figure \ref{fig:visattn} (c), we draw the scores of relationships for sorting. All the relationships are firstly sorted according to the assembling attention scores $\bm{\beta}$ induced from image caption, and then their $\bm{\eta}$ scores and the sentence likelihoods are drawn, which are used for sorting by UD-ICRC-attn and UD-ICRC respectively. The line charts show that the predicted $\bm{\eta}$ scores have a similar trend with the $\bm{\beta}$ and therefore it can correctly rank the relationships according to their importance. However, the sentence likelihoods do not show this trend, suggesting that these scores (including the product scores used in traditional scene graph) are irrelevant to the importance of the relationships. In Figure \ref{fig:visattn} (d-e), we also compare the topic linguistic scene graph and the traditional scene graph from the motifs \cite{zellers2018neural}. The topic linguistic scene graph focuses on the relationships of humans interest which are more important in the images. Besides, the scene graph of linguistic style allows the relationships to be expressed in a natural way with more suitable words, despite that the given detected object categories may be not so appropriate in the language context, \eg, in the first example, the \textit{7\_photograph} is expressed as \textit{station} in the relationships.

\subsection{Topic Scene Graph for Retrieval}
As the topic scene graph provides relationships relevant to the major events in an image, it can be utilized for image retrieval \cite{lee2018stacked,wang2020cross}. We adopt the classic image-text matching model SCAN \cite{lee2018stacked}. 1,000 images are randomly chosen from the test set, and their top 1 or 5 relationships are collected as query for retrieving correct target images. The recall (R@K, K is 1, 5, 10) and the median rank of the correctly retrieved images \cite{Kim_2019_CVPR} are used as the metrics. We run through this process 3 times and report average results. Significant improvement brought by attention alignment is observed in Table \ref{tab:retrieval}. In addition, some major events can be decomposed into multiple relationships, \eg, the major events of the query image in Figure \ref{fig:retrieval} (left column) can be expressed with two relationships which are the top two given by our topic scene graph. If one directly uses the original image or traditional scene graph to retrieve similar images, the results may be not the desired ones. The proposed topic scene graph provides fine-grained descriptions of major events and makes it possible to designate the target content to be retrieved, \eg, to retrieve \textit{woman talking on telephone} or \textit{woman sitting on bench}. 

\begin{table}[]
\setlength{\abovecaptionskip}{-0.0cm}
\caption{The image retrieval results using top 1 relationships. We use the recall at K (R@K, higher is better) and the median rank of the target image (Med, lower is better). }
\begin{center}
\resizebox{0.40\textwidth}{!}{
\begin{tabular}{@{}c|cccc@{}}
\hline
Model         & R@1           & R@5            & R@10           & Med            \\ \hline
TriLSTM       & 1.73          & 7.47           & 12.83          & 135.33         \\
UD-ICRC       & 5.67          & 20.40          & 31.73          & 27.33          \\
UD-ICRC-attn  & \textbf{9.73} & \textbf{31.67} & \textbf{46.13} & \textbf{12.33} \\ \hline
UD-ICRC-label & 17.77         & 49.17          & 67.37          & 5.67           \\ \hline
\end{tabular}}
\vspace{-0.6cm}
\end{center}

\label{tab:retrieval}
\end{table}

\begin{figure}[t]
 \setlength{\abovecaptionskip}{-0.3cm}
 \setlength{\belowcaptionskip}{-0.3cm}
\begin{center}
\includegraphics[width=1.0\linewidth]{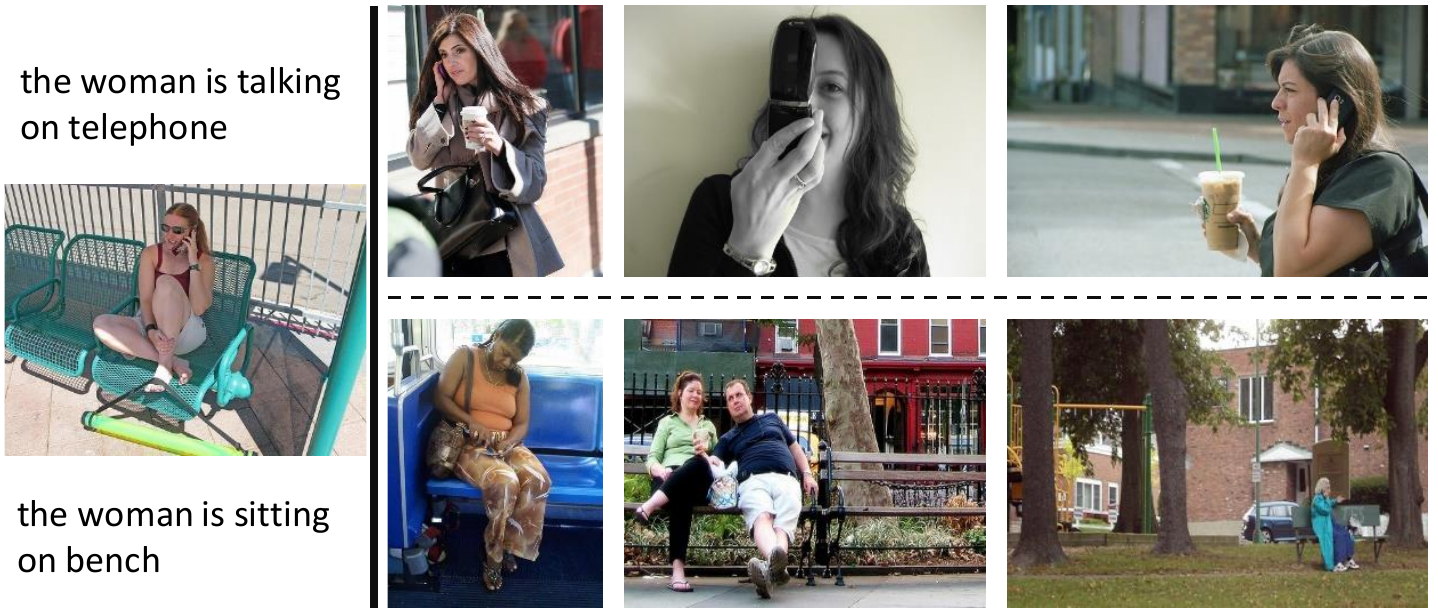}
\end{center}
   \caption{Two important relationships given by topic scene graph of the left image are used to retrieve similar images respectively. The results are shown on the right column. }
\label{fig:retrieval}
\end{figure}

\section{Conclusion}
In this work, we propose to generate the scene graph jointly with the image caption so that it can not only understand the image comprehensively, but also balance the important and trivial contents. The attention information from the image caption 
provides guideline to emphasize the important relationships. In addition, we generate the scene graph together with the image caption using a shared captioning module, making it express in a more natural style. Experimental results show the advantages of the proposed method in both performance and its feasibility in mining the important relationships without strong supervision. Besides, the topic scene graph has shown its practicality for controllable and fine-grained retrieval. 

\vspace{0.1cm}
\noindent \textbf{Acknowledgements.} This work is partially supported by National Key R\&D Program of China (2020AAA0105200), Natural Science Foundation of China under contracts Nos. U19B2036, 61922080, 61772500, and CAS Frontier Science Key Research Project No. QYZDJ-SSWJSC009.


\clearpage
{\small
\bibliographystyle{ieee_fullname}
\bibliography{egbib}
}

\end{document}